\documentclass[fleqn,10pt]{wlscirep}
\usepackage[utf8]{inputenc}
\usepackage[T1]{fontenc}

\usepackage{wrapfig}

\newcommand{\eq}{\begin{equation}}
\newcommand{\eqx}{\end{equation}}
\newcommand{\eqn}{\begin{eqnarray}}
\newcommand{\eqnx}{\end{eqnarray}}

\renewcommand{\th}{\theta}

\newcommand{\qq}{\quad\quad}

\newcommand{\gen}{\mathcal{G}}
\newcommand{\block}{\mathcal{B}}

\usepackage{tikz}

\tikzstyle{tensor}=[rectangle,draw=blue!50,fill=blue!20,thick]

\tikzstyle{tensornl}=[rectangle,draw=violet!50,fill=violet!20,thick]

\tikzstyle{tensorio}=[rectangle,draw=green!50,fill=green!20,thick]

\newcommand{\blockres}[2]{
\draw[tensor] #1 rectangle +(1, -1);
\path #1 +(0.5,-0.5) coordinate (A);
\draw (A) node {#2};
}

\newcommand{\blocknonloc}[2]{
\draw[tensornl] #1 rectangle +(1, -1);
\path #1 +(0.5,-0.5) coordinate (A);
\draw (A) node {#2};
}

\newcommand{\blockout}[2]{
\draw[tensorio] #1 rectangle +(1, -1);
\path #1 +(0.5,-0.5) coordinate (A);
\draw (A) node {#2};
}

\newcommand{\blockinp}[2]{
\draw[tensorio] #1 rectangle +(1, -2);
\path #1 +(0.5,-1) coordinate (A);
\draw (A) node {#2};
}

\newcommand{\networkfigure}{
\hspace{-1cm}\begin{tikzpicture}[inner sep=1mm,baseline={([yshift=0ex]current bounding box.center)}]

\node (0) at (-1.5, -0.5) {$\{z_i\}_{i=1}^{128}$};
\node (1) at (-1.5, -1.5) {$c$};
\node (2) at (15.75, -0.5) {$X$};

\draw[-] (0) -- (5.15, -0.5);
\draw[-] (5.85, -0.5) -- (11.65, -0.5);
\draw[->] (12.35, -0.5) -- (2);

\draw[dotted] (5.15, -0.5) -- (5.85, -0.5);
\draw[dotted] (11.65, -0.5) -- (12.35, -0.5);

\draw[-] (1) -- (5.15, -1.5);
\draw[-] (5.85, -1.5) -- (11.65, -1.5);
\draw[-] (12.35, -1.5) -- (13, -1.5);

\draw[dotted] (5.15, -1.5) -- (5.85, -1.5);
\draw[dotted] (11.65, -1.5) -- (12.35, -1.5);

\draw[->] (1.5, -1.5) -- (1.5, -1);
\draw[->] (3, -1.5) -- (3, -1);
\draw[->] (4.5, -1.5) -- (4.5, -1);
\draw[->] (6.5, -1.5) -- (6.5, -1);
\draw[->] (8, -1.5) -- (8, -1);
\draw[->] (11, -1.5) -- (11, -1);
\draw[->] (13, -1.5) -- (13, -1);

\blockinp{(-0.5, 0)}{In}

\blockres{(1.0, 0)}{$\block_1$}
\blockres{(2.5, 0)}{$\block_2$}
\blockres{(4, 0)}{$\block_3$}

\blockres{(6.0, 0)}{$\block_7$}
\blockres{(7.5, 0)}{$\block_8$}
\blocknonloc{(9.0, 0)}{$\block_9$}
\blockres{(10.5, 0)}{$\block_{10}$}
\blockres{(12.5, 0)}{$\block_{13}$}

\blockout{(14.0, 0)}{Out}

\end{tikzpicture}
}

\title{Aesthetics and neural network image representations}

\author[1,*]{Romuald A. Janik}
\affil[1]{Jagiellonian University,
Institute of Theoretical Physics and Mark Kac Center for Complex Systems Research,
ul. {\L}ojasiewicza 11,
30-348 Krak{\'o}w,
Poland}

\affil[*]{romuald.janik@gmail.com}

\begin{abstract}
We analyze the spaces of images encoded by generative neural networks of the BigGAN architecture.
We find that generic multiplicative perturbations of neural network parameters away from the photo-realistic point often lead to networks generating images which appear as \emph{``artistic renditions''} of the corresponding objects. This demonstrates an emergence of aesthetic properties directly from the structure of the photo-realistic visual environment as encoded in its neural network parametrization.
Moreover, modifying a deep semantic part of the neural network leads to the appearance of symbolic visual representations.
None of the considered networks had any access to images of human-made art.
\end{abstract}
\begin{document}

\flushbottom
\maketitle
%
%
\thispagestyle{empty}

\section*{Introduction}

\begin{wrapfigure}{r}{0.5\textwidth}
\begin{center}
  \includegraphics[width=0.48\textwidth]{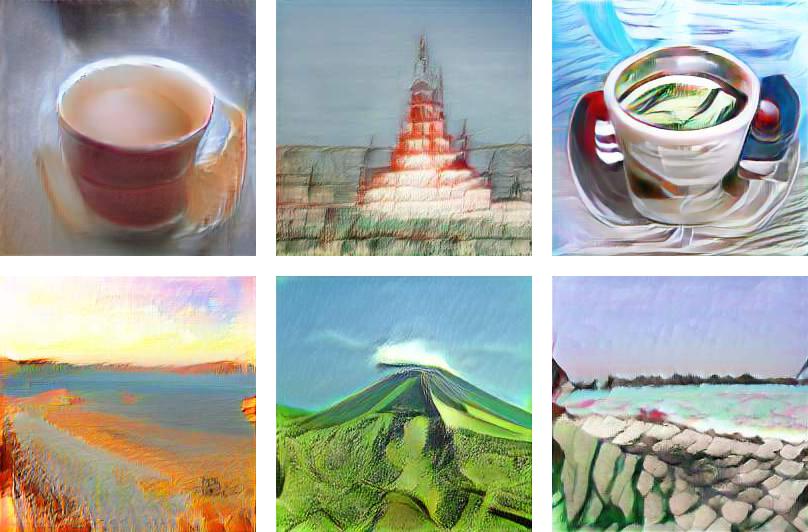}
  \end{center}
  \caption{Selected images generated by neural networks obtained through  various ways of randomized modifications from a BigGAN network generating photo-realistic images\\ (for further examples see \url{https://neuromorphic.art}).}
  \label{fig.sample}
\end{wrapfigure}

Among the many strands of contemporary Machine Learning, a prominent place is taken by generative neural networks~\cite{GAN,GANreview,VAE,WAE}. These neural networks aim to generate new, unseen examples based on a given dataset and thus aim to learn the variability structure of the data. Of particular interest for the present investigation are neural networks which generate photo-realistic images of the natural and human environment. In order to do so, they have to incorporate extensive knowledge about the structure of the visual photo-realistic world. This information is encoded in a nontrivial way in the weights of the neural network layers. The goal of this work is to explore global properties of this neural encoding. To the best of our knowledge, such properties have not been investigated so far.


In the present paper, we show two surprising features of these neural encodings.
First, moving away from the photo-realistic world in a generic (multiplicative) manner leads in many cases to the emergence of \emph{``artistic rendition''} and aesthetic properties as perceived by humans.
Second, upsetting a part of deep semantic information leads to the appearance of imagery which can be interpreted as \emph{symbolic visual representations}.
These results may have far reaching interdisciplinary consequences, touching upon
our understanding of the neural basis of aesthetics (neuroaesthetics)~\cite{Zeki1,Zeki2,Ramachandram,Nadal,Chatterjee,LiZhang}, the theory and philosophy of art and, given the similarities between deep convolutional networks and the visual cortex~\cite{VisualCNN1, VisualCNN2, VisualCNN3}, these results may inspire novel investigations within cognitive neuroscience.


The two main classes of generative neural networks are 
Generative Adversarial Networks (GAN)~\cite{GAN} which appear in a multitude of variants~\cite{GANreview}, as well as
Generative Autoencoders e.g. Variational Autoencoders (VAE)~\cite{VAE}, Wasserstein Autoencoders (WAE)~\cite{WAE} and many others. 
At the end of training these constructions provide for us a~\emph{generator network} $\gen_\th$ with a given set of ``optimal'' weights (i.e. neural network parameters)  $\th=\th_*$. The resulting network generates an image $X$
\eq
\label{e.generator}
X = \gen_{\th=\th_*} (\{z_i\}, c)
\eqx
given as input a set of \emph{latent variables} $\{z_i\}$, and optionally (depending on the particular construction) the class $c$ to which the generated image should belong. The latent variables $\{z_i\}$ are usually drawn from some random distribution and encode the variability of the overall space of images. 

Most of the focus in this domain of Machine Learning research is concentrated on
finding the optimal architecture and training procedure so that the generated images represent best the space of images given as training data. 
In the present paper, we would like to pursue, however, an orthogonal line of investigation and study how the generated space of images changes as we move in the space of neural network parameters.

Indeed, a particular generator neural network $\gen_{\th=\th_*}$ with the specific set of optimal weights~$\th_*$, obtained through training 
on the \emph{ImageNet} dataset~\cite{imagenet},  may be understood as providing a \emph{neural network representation} of the space of photo-realistic natural images (including also human-made structures, objects, vehicles, food etc. but no art).

Similarly, we can view each particular choice of parameters $\th$ of the generator neural network $\gen_\th$ as encoding some specific \emph{space of images}. We thus have a mapping
\eq
\th \quad \longrightarrow \quad \text{\textit{space of images generated by $\gen_\th$}}
\eqx
The above mentioned optimal weights
$\th_*$ get mapped to the space of photo-realistic natural images.
The goal of this paper is to investigate in what way the \emph{space of images} changes as we move away from the photo-realistic point~${\th=\th_*}$ in the space of neural network parameters.

\section*{Results}

\subsection*{Aesthetics and artistic rendition}

\begin{figure*}
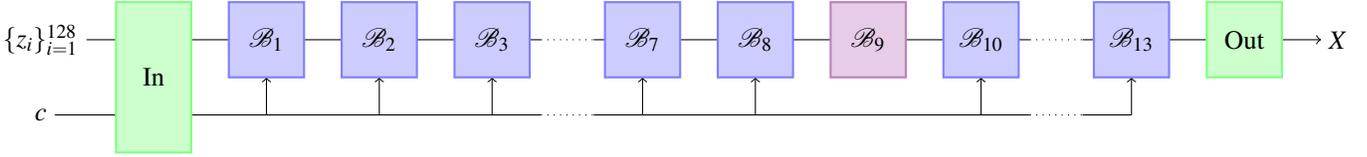

  \hspace{0.75cm}\networkfigure
  \vspace{0.25cm}
  \caption{Schematic structure of the \texttt{BigGAN-deep-256} generator network $\gen_\th$ taking as input a 128-dimensional vector of latent variables $\{z_i\}$ and one of 1000 ImageNet classes $c$. The blue blocks are residual blocks with two $1\times 1$ and two $3\times 3$ convolutions as well as four conditional batch normalization layers which receive shortcut connections from the entry stage. The purple block is a ``self-attention'' block. 
  The blocks $\block_2$, $\block_4$, $\block_6$, $\block_8$, $\block_{11}$ and $\block_{13}$ increase image dimensionality by factors of 2.
  See~\cite{BigGAN} for details.
  }
  \label{fig.network}
\end{figure*}

For the experiments performed in the present paper we utilized the generator part of the \texttt{BigGAN-deep-256} network~\cite{BigGAN} trained on the \emph{ImageNet} dataset~\cite{imagenet}.
The general structure of the network is shown in Fig.~\ref{fig.network}.
It consists of an entry stage, followed by 13 groups of layers, which also receive shortcut connections directly from the entry stage, followed by the output stage (see the original paper~\cite{BigGAN} for details). The generator network has around 55~million trainable parameters~$\th$. As the photo-realistic point, we take the weights $\th_*$ of the pretrained model~\cite{BigGANmodel}.

In order to move away from $\th_*$, we employ a \emph{multiplicative} random perturbation of the neural network parameters
\eq
\label{e.perturb}
\th = \th_* \cdot \left( 1 + \alpha \cdot \text{\sl random} \right)
\eqx
where $\cdot$ denotes element-wise multiplication,
{\sl random} is of the same shape as $\th_*$ and is drawn from a normal distribution with zero mean and unit standard deviation. 
We take the constant $\alpha=0.35$ so that we move noticeably away from the photo-realistic point~$\th_*$. The precise value of $\alpha$ does not matter much.

It is important to contrast here moving in latent space $\{z_i\}$ for a fixed generative neural network, which is often studied in the Machine Learning literature, with moving in the space of weights $\th$, which we do in \eqref{e.perturb}. 
In the former case, each point $\{z_i\}$ corresponds to a single image generated by the fixed generator network $\gen_{\th_*}$. Thus, when varying $\{z_i\}$, one moves in the given fixed photo-realistic space of images associated to $\gen_{\th_*}$.
In contrast, in the case studied here and given by \eqref{e.perturb},
each point $\th$ is a \emph{different} generator network $\gen_\th$, and thus corresponds to a different \emph{visual universe} of images which can be potentially generated by~$\gen_\th$.

\begin{figure*}[t]
  \includegraphics[width=\textwidth]{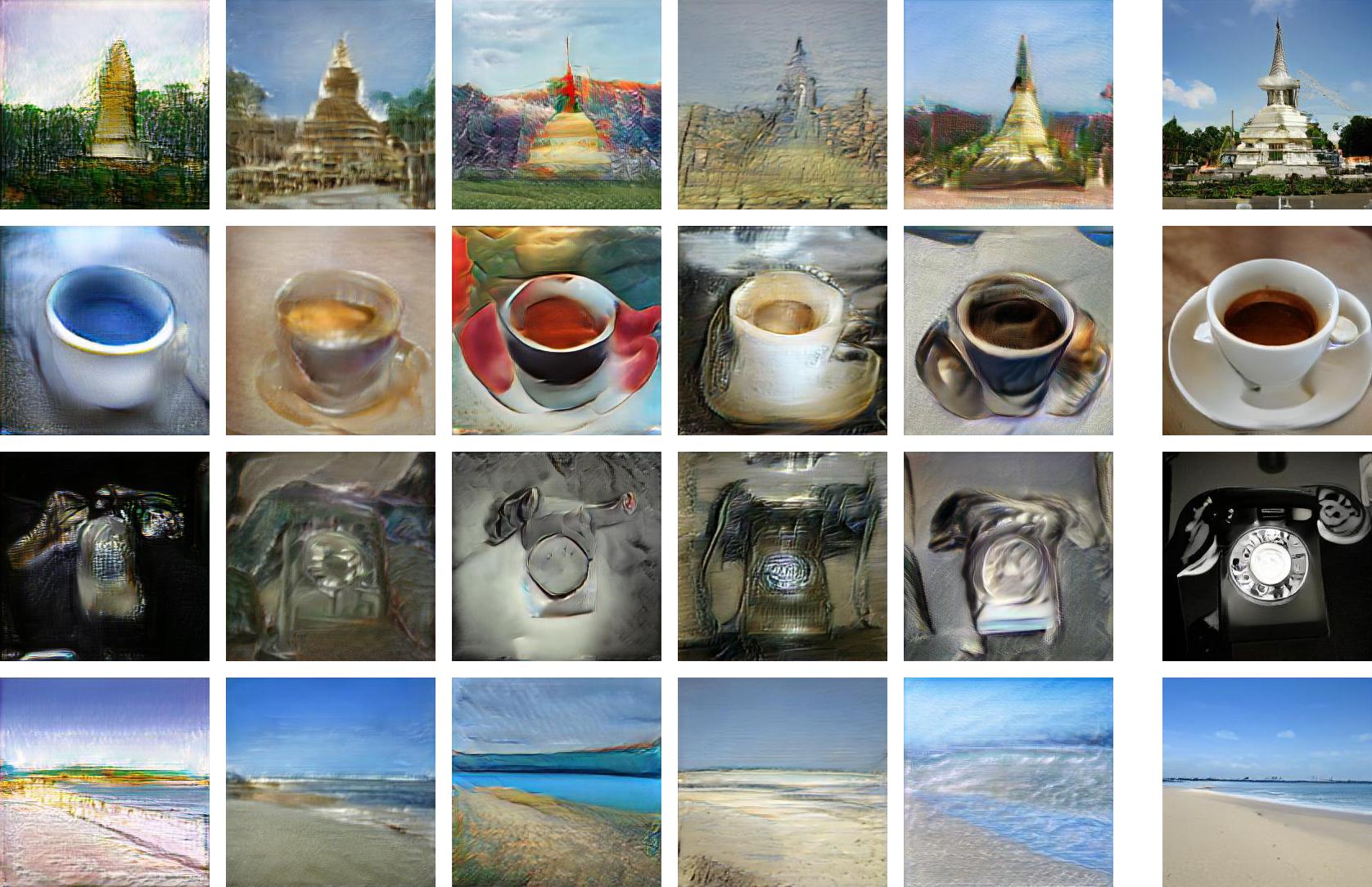}
  \caption{Images generated by neural networks with weights given by (\ref{e.perturb}), realizing various deviations from the photo-realistic point $\th=\th_*$. Each column corresponds to a distinct neural network. None of the networks had access to any human-made art. Far right: corresponding photo-realistic images generated by the original \texttt{BigGAN-deep-256} network.
  The inputs to the different networks were identical.}
  \label{fig.perturb}
\end{figure*}

In Fig.~\ref{fig.perturb} we show images generated by five networks $\gen_\th$ with weights $\th$ given by random perturbations of the form~\eqref{e.perturb} with random seeds chosen from the range $0-10$ and, for comparison, images produced by the original photo-realistic network $\gen_{\th=\th_*}$. The images represent \textit{stupa}, \textit{espresso}, \textit{dial telephone} and \textit{seashore}.
The respective latent noise $\{z_i\}$ inputs in each row of Fig.~\ref{fig.perturb} for all networks were identical. The two leftmost images in the top row\footnote{The remaining four images in Fig.~\ref{fig.sample} were obtained by substituting a specific \emph{subset} of weights by randomly drawn values.} of Fig.~\ref{fig.sample} were also obtained using~\eqref{e.perturb}.

A striking feature of the obtained images is that they seem to give an \emph{``artistic rendition''} of the original photo-realistic objects.
The perturbation of the space of parameters~$\th$ away from the point $\th_*$ clearly breaks the fine-tuning necessary for the photo-realistic rendition of the images by the original network, which is of course not surprising.
What is quite unexpected, however, is that this manner of breaking leads to aesthetically pleasing and interesting images, at least for a range of object classes. 
The deformations of photo-realism are reminiscent of the kind of simplifications that a human artist would employ when painting or making a rough sketch. 
Indeed, many of the obtained images could arguably be mistaken at first glance for paintings or sketches made by a human.
Moreover, in the majority of cases the utilized colour palette and colour transitions appear balanced and aesthetic --- they do not strike us as artificial, which would be a natural expectation given the random character of the perturbation~\eqref{e.perturb} of the photo-realistic network parameters $\th_*$. Most probably, the multiplicative character of the perturbation~\eqref{e.perturb} helps in this respect as small weights do not get disproportionally large modifications.

Another intriguing feature is that 
quite often one can discern a particular style characteristic of a specific perturbed neural network, which differentiates it from neural networks obtained through other perturbations. This can be seen as a certain visual consistency in the columns of Fig.~\ref{fig.perturb}.




Let us also mention some limitations of these results, which have to be kept in mind. Only a subset of \emph{ImageNet} classes (architectural, some objects, landscapes) behaves equally well under these deformations. Most probably the other ones require more fine-tuned weights. Indeed, generated images for certain classes exhibit some pathologies even at the photo-realistic point~$\th_*$, i.e. for the original pretrained network. Imposing further weight perturbations in these cases may easily ``break'' the images.
Furthermore, we do not claim that \emph{every} randomized perturbation leads to an aesthetic result. However, quite a lot do (in particular, the examples shown in Fig.~\ref{fig.perturb} were chosen just out of consecutive random seeds 0-10).
As a further test of genericity, in the Supplementary Fig. S1, we show examples of a \emph{stupa} and \emph{espresso} for a wide range of \emph{consecutive} random seeds, without any human selection.
Let us note that
occasionally one encounters visually stunning examples such as those shown\footnote{The two first images in the top row of Fig.~\ref{fig.sample} were obtained using \eqref{e.perturb}.} in Fig.~\ref{fig.sample}. One might make here an analogy with the wide spectrum of artistic talent in the human population.

Finally, we would like to emphasize that
the appearance of aesthetic properties should be interpreted in the context of the immense dimensionality of the space of parameters (${\sim 55\; million}$). 
In such a high dimensional space any two randomly chosen directions of deformation are essentially mutually orthogonal. Therefore, if any qualitative property \emph{repeats itself under random sampling} even in a subset of cases, it is, in our opinion, a~significant observation and the aforementioned property can be considered as being to a large degree generic.


\subsubsection*{What do the above experiments tell us?}

Firstly, we may infer that the property of being perceived as aesthetic by a human may be related to the very nature of the photo-realistic world. 
Indeed, the original network was exposed only to photo-realistic images and did not have any contact with human art. The perturbations of the neural parametrization of the space of images leading to the ones shown in Fig.~\ref{fig.perturb} were generic (multiplicative) random perturbations which were not biased by any further image input or optimization procedure.

Secondly, this property is firmly tied to the \emph{neural network representation} $\gen_\th$ of spaces of images of the human visual environment, which apart from the particular values of the parameters $\th$, incorporates as a kind of structural prior the specific generator architecture of the \texttt{BigGAN-deep-256} model.

Thirdly, the observed interplay of aesthetics and neural parametrization ties in with the hypothesis of~\cite{Zeki1, Zeki2, Ramachandram} that the perception of aesthetics is linked with features of the human visual system in the brain. 
This may go beyond being just an analogy as
there are already indications that the higher stages of human visual processing are quite well correlated with deeper levels of convolutional neural networks~\cite{VisualCNN1,VisualCNN2,VisualCNN3}. We will return to this point in more detail in the \emph{Discussion} section.

Finally, we believe that the above findings could be of potential interest for humanities, in particular for the theory and philosophy of art and aesthetics. In this respect, the results of the experiments performed in the present paper could be treated as providing an unexpected piece of evidence
of a latent possibility
of a biological (non-cultural) origin of some kinds of ``artistic renditions''.

\subsubsection*{Differences with other approaches}

It is important to contrast the results obtained in the present paper with some other approaches linking artistic renditions and neural networks as superficially they may seem similar. 

A~very well known construction is the so-called \emph{Neural Style Transfer}~\cite{NeuralStyle}, where a given input image is transformed into the style of a second image (the style image), typically an image of a painting or work of art, with the similarity in style measured by a deep neural network pretrained on an image classification task.
Alternatively, GANs have also been trained on art (see e.g.~\cite{GANArt}) to generate new images based on the given artistic styles.
These techniques use explicit input of human-made art to produce new images similar in style, which was of course their key goal. 

The aim of our investigation was, however, quite different and the ``artistic character'' of the images described in the present paper appeared spontaneously as an \textit{a-priori} unexpected byproduct.
Our results were obtained using only photo-realistic images of the natural world and the human environment without any contact with human or machine-made art. They thus provide a realization of the \emph{emergence} of aesthetic properties directly from a neural network parametrization of a  photo-realistic world.


\begin{figure*}[t!]
  \includegraphics[width=\textwidth]{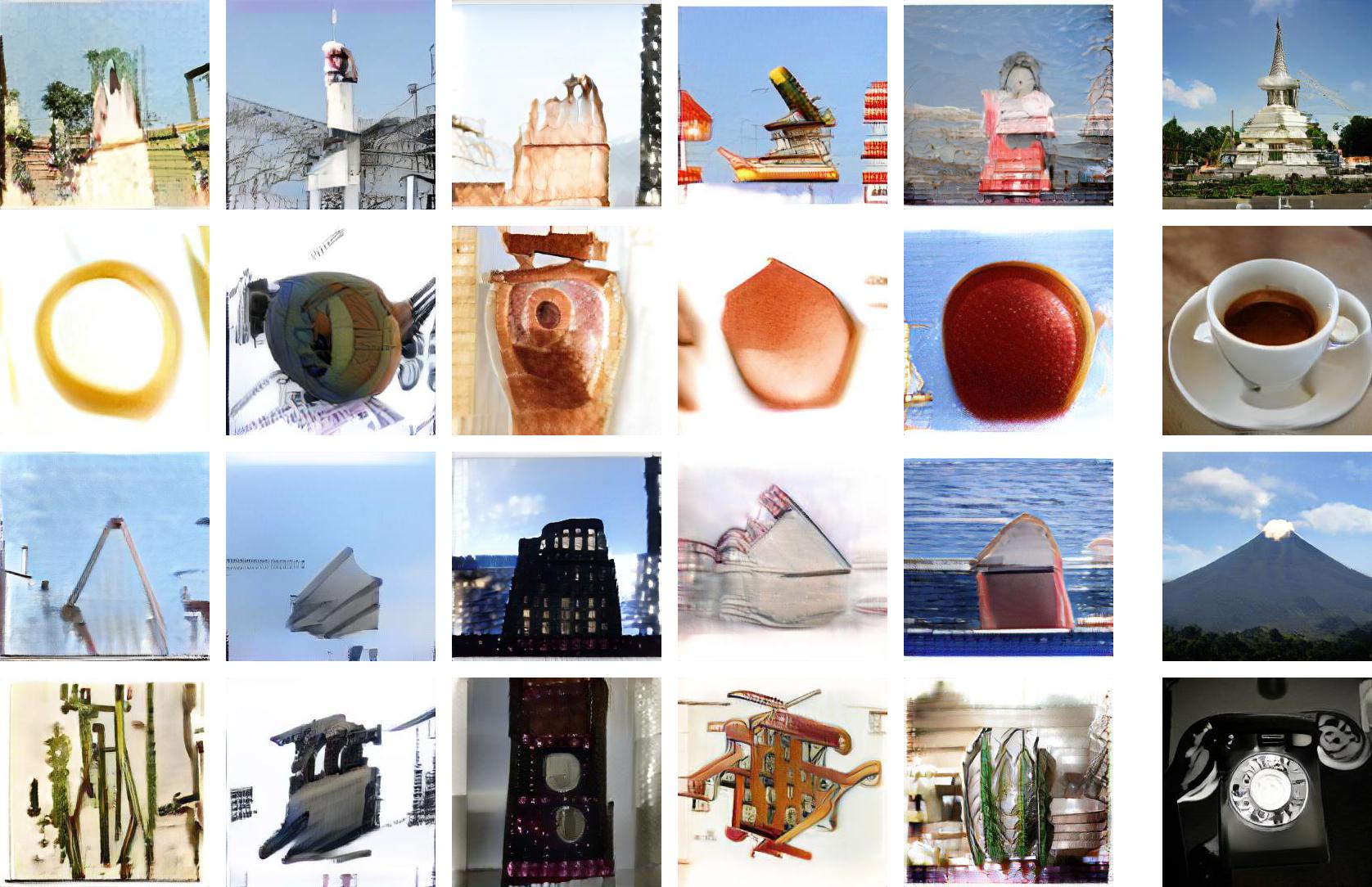}
  \caption{Images  generated  by  neural  networks  with  weights  given  by~\eqref{e.random},  upsetting  the deep semantic structure of the representation of the space of images.
  Each column corresponds to a distinct neural network.
  Most of the images exhibit dominant features of the original object realized in terms of different ingredients (see text).
  Far right:  corresponding photo-realistic images generated by the original \texttt{BigGAN-deep-256} network. The inputs to the different networks were identical.}
  \label{fig.symbolic}
\end{figure*}

An approach perhaps closest in spirit to ours is the hand picking of \emph{exceptional} latent variables $\{z_i\}$ for the photo-realistic model~$\th_*$ in order to generate surreal images (see~\cite{nnartreview} for a discussion).
That procedure really exploits the \emph{deficiencies} of the generative photo-realistic model in order to produce artistically interesting images.
The more modern text-to-image generators DALL-E and DALL-E2 from OpenAI~\cite{DALLE} can generate stunning images especially through paradoxical input text (which can be thought of as an analog of the exceptional latent variables $\{z_i\}$ mentioned above), hence receive essential input from a human. Moreover, their training data is much richer and includes, in particular, human art. 
Our result is conceptually quite different, as we show the essentially \emph{generic} appearance of aesthetic/``artistic'' images in the neighbourhood of the photo-realistic point $\th_*$, without any human intervention (see e.g. also Supplementary Fig. S1) and without any contact with human art.

\subsection*{Visual symbolic representations}

Another surprising feature of the generative neural representation of the space of images provided by the \texttt{BigGAN-deep-256} model is that it allows to exhibit a certain kind of visual symbolic representations. 
In order to see that, we first heuristically identify the location of some high level semantic information in the neural encoding of $\gen_\th$.

In contrast to the usual neural networks used for classification, the flow of information in a generative network generally runs from the most semantic/global features (incorporating here the class $c$ given as input) to the low-level pixel-based visual output.
We may therefore expect, that closer to the input we have more high-level semantic information.

In a subsequent experiment, we substitute the parameters of the second block of layers,
which we denote by
$\block_2$ (see Fig.~\ref{fig.network}), with values drawn randomly from normal distributions:
\begin{align}
\th_L &= random_L \qq & L \in \block_2 \nonumber \\ 
\th_L &= \th_{*L} \qq & L \notin \block_2
\label{e.random}
\end{align}
The parameters of the normal distributions are taken from the statistics of $\th_*$ for the corresponding weights\footnote{For convolutional weights we computed the statistics for each pixel of the $3\times 3$ filter individually. This last choice is, however, really inessential for the phenomenon discussed here.} of the given layer~$L \in \block_2$.
This can be viewed as upsetting only a deep semantic part of the neural representation of the space of images. We will comment on the specific choice of the second block $\block_2$ further below.

Images generated by neural networks $\gen_\th$  constructed through~\eqref{e.random} for five random seeds in the range $0-10$ are shown\footnote{In the Supplementary Fig. S2, we show more examples of a \emph{stupa} and an \emph{espresso} for a wider range of \emph{consecutive} random seeds in order to assess the genericity of the discussed phenomenon.} in Fig.~\ref{fig.symbolic}. 
At first glance, individually the images may seem haphazard and quite disconnected from the original photo-realistic objects, but viewing them side by side we observe surprising similarities. 
Indeed, an overall distinctive characteristic of the original object seems preserved -- like the round shape of the \emph{espresso} or the triangular form of the \emph{volcano}. It is however articulated using quite different and varying graphics primitives and materials. A similar phenomenon, but on a slightly more subtle level, occurs also for the \emph{stupa} in the first row. There, the overall shape morphs either into some quasi-architectural form or into a person-like depiction. The \emph{dial telephone} in the bottom row is most extreme. Here one cannot really identify \emph{by eye} a strong dominant feature, so its visual representations may be difficult to interpret -- although one could perhaps put forward some arguments for certain specific networks.


The above deformations can be thus understood as inducing a \emph{visual symbolic representation}, where a dominant strong characteristic of the original object 
is realized in terms of completely unrelated materials and ingredients\footnote{For strawberries (not shown here), on the other hand, it is the material -- red texture -- which seems to be the dominating feature.} 
We expect this interpretation to hold under the condition that such a very strong simple dominant feature exists for the object class in question.

The fact that the dominant visually prominent feature is still present after the modification of the weights in \eqref{e.random}, indicates that it must be encoded also in the undeformed parts of the network. From this point of view, the second block of layers $\block_2$ seems to play a privileged role in the neural network representation $\gen_\th$, as it does not destroy that feature but rather swaps in varying local visual ingredients while still preserving some sharpness and locally detailed depiction.

Upsetting similarly only the first block $\block_1$ loses any resemblance to the original objects, while doing the same for further blocks leads first to a loss of the \emph{local} photo-realistic depiction and sharpness still seen in Fig.~\ref{fig.symbolic}, while for still further blocks the original object becomes more and more directly recognizable.  
Consequently, from the point of view of the visual symbolic representations, the block $\block_2$ is essentially singled out. 

The fact that the phenomenon is mostly restricted to a subset of the neural network architecture should not be understood as a problem. First, we do not expect that all blocks/layers in a deep neural network play an equivalent role. The differentiation of their roles is in fact a very interesting feature (recall that e.g. the visual system in the brain has clear non-interchangeable modular structure). 
Second, the subset is quite sizeable, as the dimensionality of the $\block_2$ parameter space in~\eqref{e.random} is still very large, equal to around 8.5 million.
Last but not least, we find it extremely intriguing that examples of such visual symbolic representations can be indeed realized in an artificial neural network context.

\section*{Discussion}

In this paper we studied the global properties of a generative neural network parametrization of \emph{spaces of images}. We found that essentially generic deviations of the neural network parameters from the photo-realistic point~$\th_*$ quite often lead to neural networks which generate images which may appear aesthetic to humans. In many cases these images are difficult to distinguish at first glance from images of  paintings or sketches made by a human, even though the neural network did not encounter any human-made art. 

The above observation shows that aesthetic properties could arise in an \emph{emergent} way directly from the nature of the photo-realistic visual environment through the character of its neural network representation.
What is particularly intriguing about this result 
arises from tension with the belief 
that aesthetic perception is intimately linked to the human observer and appears to us as very subjective. Yet, the artificial neural network construction presented in this paper in some sense \emph{objectivizes} this quality. This opens up numerous questions. 
What is the interplay between subjectivity and objectivity in aesthetic perception?
To what extent and at what level can we draw an analogy between aspects of the neural parametrization and the biological roots of aesthetic perception in the human brain~\cite{Nadal, Chatterjee, LiZhang}?
In particular, how does this fit with the hypothesis of~\cite{Zeki1, Zeki2, Ramachandram} that aesthetic perception is related to the human visual system in the brain? 

On the one hand, as already mentioned, there is research showing that activations in the human visual cortex as measured by fMRI are quite well correlated with features in a deep convolutional network~\cite{VisualCNN1,VisualCNN2,VisualCNN3}.
Also the RGB encoding used in images being input to the artificial neural networks already takes into account some very elementary aspects of human color vision.

On the other hand, the cited results~\cite{VisualCNN1,VisualCNN2,VisualCNN3} were obtained for discriminative/classification networks and, as far as we know, there is no similar investigation for generative networks. 
Indeed in the latter case, 
the information flow goes in an opposite direction as the generative networks produce images (thus intuitively mimicking visual imagination), while the human brain in the studies~\cite{VisualCNN1,VisualCNN2,VisualCNN3} perceives them. 
Of course, the human brain visual system with bidirectional information flow is certainly quite different in detail from a standard feedforward convolutional neural network. Paradoxically, this may increase the potential relevance of generative neural networks as they may be considered as modelling the top-down pathway in perception (e.g. along the lines of \cite{LEEMUMFORD}).
In addition, one should note that there are marked similarities between perception and visual imagery seen in neuroimaging studies~\cite{DIJKSTRA} (see also~\cite{ALBRIGHT} for an extended discussion).

On a higher, more qualitative level, a~common feature of the analysis of aesthetic perception motivated by neuroscience in~\cite{Zeki1, Zeki2, Ramachandram} is its emphasis on the essences of particular concepts, characteristic of the brain seeking \emph{constancy} in its environment and thus abstracting away transient particularities. 
From this point of view, a pictorial representation which is closer to the internalized essence is more likely to be perceived as aesthetic. 
Photo-realistic details, on the other hand, are specific to particular object instances and tend to lower the aesthetic appeal. 

In this sense, one can view the randomized perturbations away from the photo-realistic point as dispensing with the fine-tuned particular details, which due to the uncorrelated nature of the perturbations would get averaged out.
The outcome could thus be interpreted indeed as generating more \emph{essence-like} depictions.
But this is certainly not the whole story, as just performing gaussian blurring on images 
does not make them aesthetic or \emph{essence-like}. The neural network randomized perturbations must therefore act in a more subtle way
and the concrete form of the generator neural network parametrization $\gen_\th$ somehow manages to capture some finer aspects of human aesthetic perception.
Indeed, the emergence of visually appealing forms and colour transitions probably depends crucially on properties of convolutions appearing in the neural encoding, the overall colour structure of the natural environment and the specific mixing induced by the randomized modification of weights.

In addition, the specific randomized perturbations leave an
imprint on the overall style of images generated by a particular deformed network.
One could think of this as an analog of inter-subject ``artistic''  variability.


In this respect, it is interesting to speculate to what extent natural randomness and stochasticity in the nervous system~\cite{RANDOMNESS1, RANDOMNESS2} 
could be relevant in the context of the present observations. One could expect that randomness would lead to more robust (\emph{essence-like?}) concepts. Indeed, in the artificial neural network context it has been shown that adding random noise to neural networks during training and evaluation increases their resilience to adversarial examples~\cite{ADVRND1, ADVRND2}.
This type of randomness, however, would be associated with intra-subject (or here intra-network) variability and is not directly represented in the constructions of the present paper.


The second main result of this paper is that randomly scrambling a specific part of the deep semantic structure of the neural network parameters $\th_*$ can lead to \emph{visual symbolic representations}, where a dominant visual feature of a particular object is realized in terms of atypical and nonstandard visual ingredients. 

This result is quite intriguing, as
symbolic representations are an important component appearing throughout human culture, ranging from a key element of artistic expression (see e.g.~\cite{SYMBOLS1,SYMBOLS2}) to the way that psychoanalysis interprets dreams~\cite{FREUD,JUNG}, with some important psychological concepts manifesting themselves encoded in various proxy objects, persons or scenes.
In this context, we should nevertheless emphasize that
the type of symbolic representation appearing in the present work is very much simplified, restricted just to some visual characteristics and completely blind to any aspect of cultural meaning, as the original generative neural network's world was just the purely visual environment.
Even with these caveats, however, we find it very surprising that an analog of a symbolic representation can arise naturally in an artificial neural network context.

As a side remark, let us note that all the constructions in the present paper involve various kinds of randomized ``rewirings'' of the connection strengths of the artificial neural network. If one would look for brain states where randomness is enhanced, then a natural example would be the psychedelic state, where increased neural signal diversity was measured~\cite{psychedelic1} in accordance with the ``entropic brain'' picture~\cite{psychedelic2a,psychedelic2b}. Perhaps some analogies could be pursued in this direction.


Finally, we would also like to make a methodological comment.
The method of analysis of the neural network encoding used in the above case is in fact akin to the classical practice in neuroscience/neurology of analyzing the cognitive characteristics of patients with various brain lesions as a window on the functioning of the corresponding subsystems of the brain. In the present paper, we basically artificially induced a lesion in the generator network by substituting the values of a subset of neural network weights with completely random numbers. Subsequently, we examined the resulting neural network output. 
We expect that this technique may be quite useful for analyzing the structure of deep neural network knowledge representations for very complex models.
Although here our focus was slightly different, as we emphasized more the qualitatively novel ``positive''  aspects (the visual symbolic representations) rather than the breakdown of photo-realism.


We believe that the obtained results and the consequent questions could foster new research on the borderline of cognitive neuroscience, (neuro)aesthetics and artificial neural networks. Moreover, we hope that both of the two main results of the present paper would be of potential interest for humanities, wherein they can be considered as \emph{proofs of concept} showing the possible roots of some key human phenomena.

\section*{Acknowledgements}

I would like to thank Anka Janik, Maciej A. Nowak, Igor Podolak for interesting discussions and comments and Daniel Wójcik for explanations and references on noise in the nervous system.
This work was supported by the research project \textit{Bio-inspired artificial neural networks} (grant no. POIR.04.04.00-00-14DE/18-00) within the Team-Net program of the Foundation for Polish Science co-financed by the European Union under the European Regional Development Fund and by a grant from the Priority Research Area DigiWorld under the Strategic Programme Excellence Initiative at Jagiellonian University.

\pagebreak

\appendix

\renewcommand{\thefigure}{S\arabic{figure}}
\setcounter{figure}{0}

\section{Supplementary information}

The figures S1 and S2 shown on the following pages are analogues of Fig. 3 and Fig. 4 but obtained for neural networks constructed using a \emph{consecutive} range of random seeds 0-23 used for generating either the multiplicative random perturbations or the weights in block $\mathcal{B}_2$. In this way one can assess the genericity of the results discussed in the main text. In order to fit all the~24 neural networks, only two types of images are shown: a \emph{stupa} and an \emph{espresso}.

\begin{figure}
\centering
\includegraphics[width=0.93\textwidth]{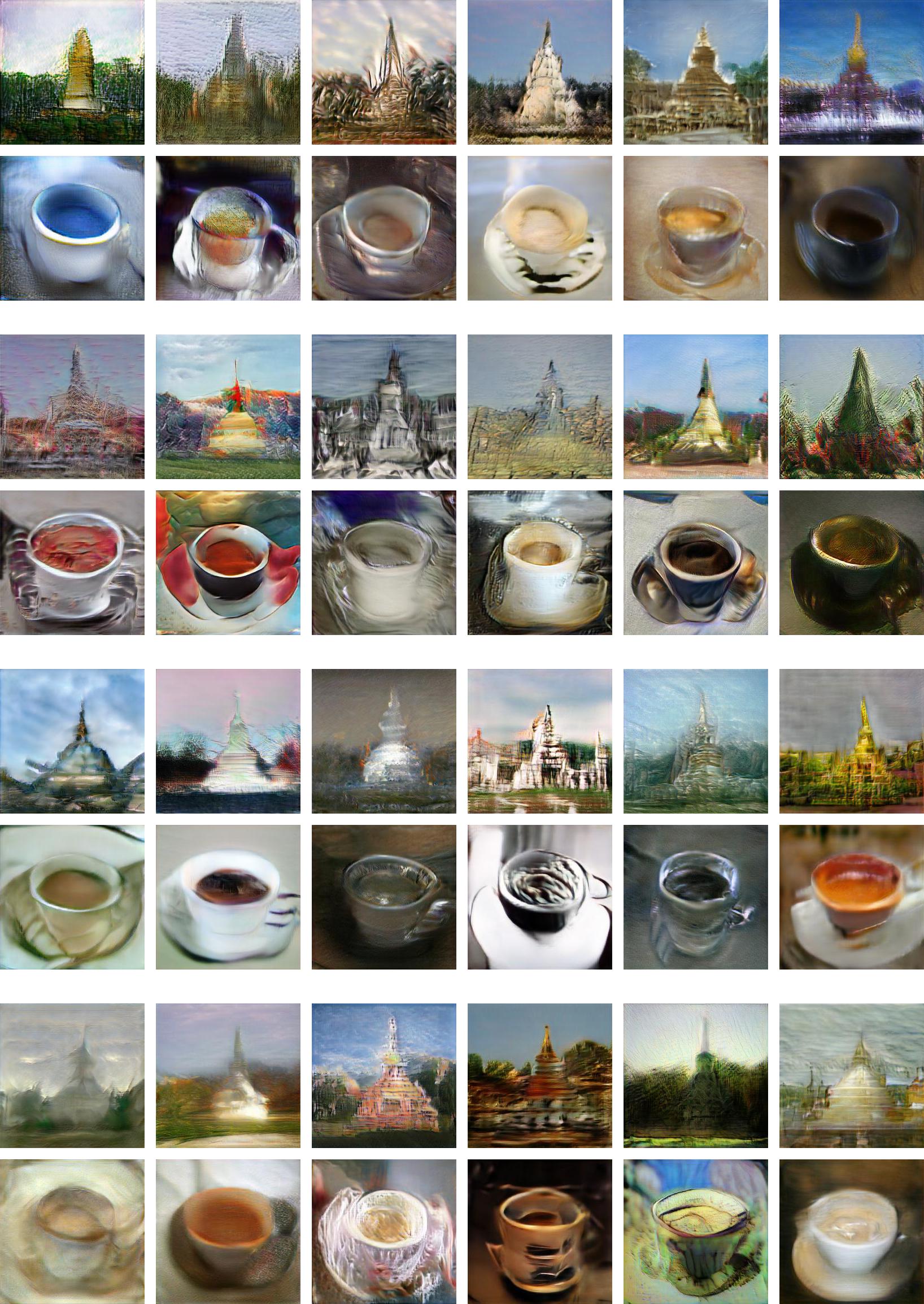}
\caption{Images of a \emph{stupa} and an \emph{espresso} generated by neural networks with weights obtained through random multiplicative perturbations (Eq. (3) in the main text) for consecutive random seeds~0-23.}
\end{figure}

\begin{figure}
\centering
\includegraphics[width=0.93\textwidth]{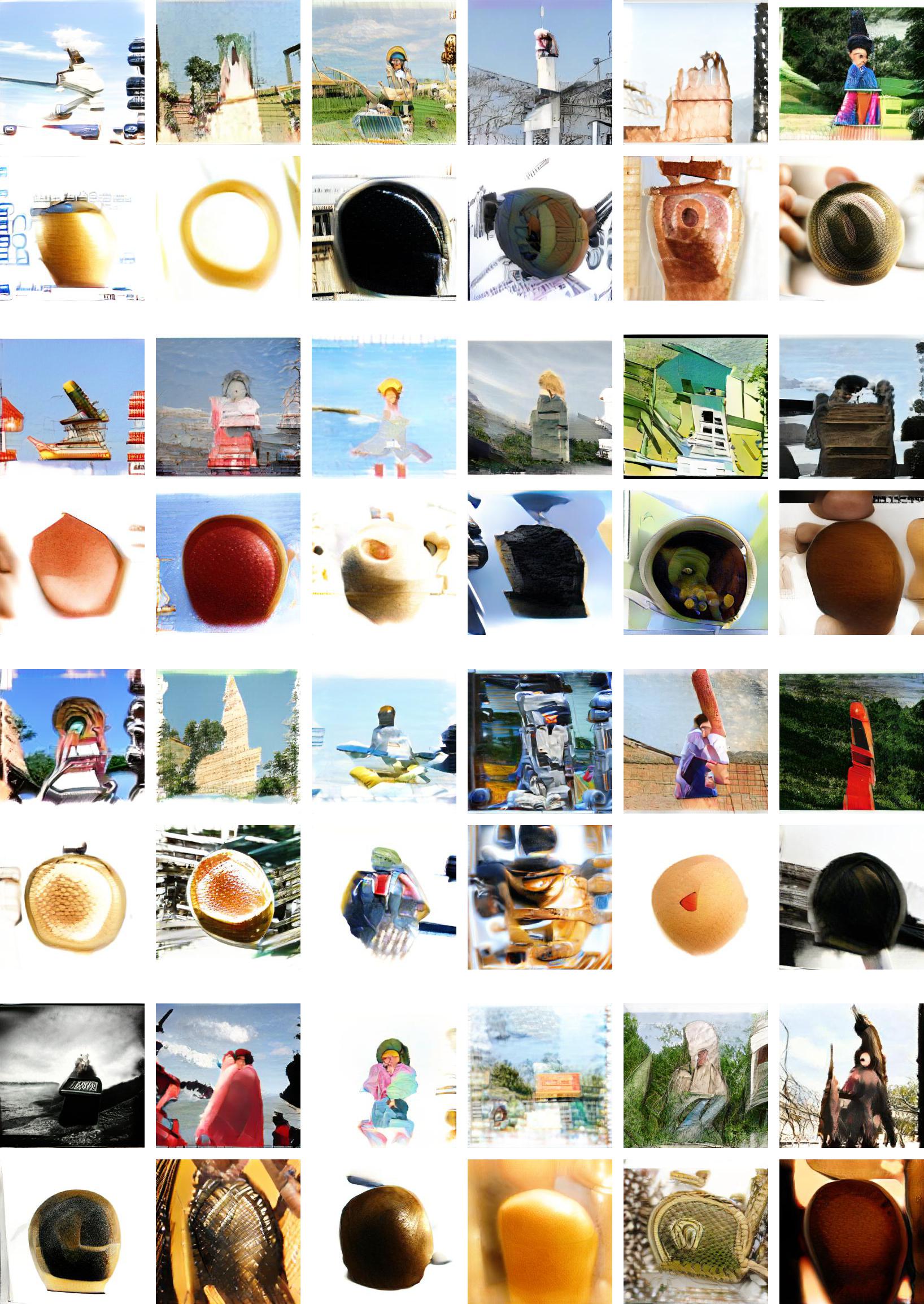}
\caption{Images of a \emph{stupa} and an \emph{espresso} generated by neural networks with weights obtained through randomly scrambling the block $\mathcal{B}_2$ (Eq. (4) in the main text) for consecutive random seeds~0-23.}
\end{figure}

\end{document}